\documentclass[10pt,twocolumn,letterpaper]{article}

\usepackage{cvpr}

\usepackage[utf8]{inputenc}
\usepackage[T1]{fontenc}
\usepackage[english]{babel}
\usepackage{amsfonts}
\usepackage{booktabs}
\usepackage{hyperref}
\usepackage{cleveref}
\usepackage{algorithm}
\usepackage{algorithmic}
\usepackage{xcolor}
\usepackage{enumitem}
\usepackage{caption}
\usepackage{placeins}
\usepackage{float}
\usepackage{afterpage}
\usepackage{booktabs}
\usepackage{placeins}

\newcommand{\gmmd}{\textsc{Gram-MMD}}
\newcommand{\gmmdshort}{\textsc{GMMD}}
\newcommand{\reals}{\mathbb{R}}

\newcommand{\placeholder}[2]{%
  \rule{0pt}{0pt}%
  \fbox{\begin{minipage}[c][#1][c]{\dimexpr\linewidth-2\fboxsep-2\fboxrule}%
    \centering\small\sffamily\color{gray} #2%
  \end{minipage}}%
}

\renewenvironment{abstract}{%
  \centerline{\large\bfseries Abstract}%
  \vspace{0.3ex}%
  \begin{quote}\small}%
  {\par\end{quote}\vskip 0.5ex}

\title{Gram-MMD: A Texture-Aware Metric for Image Realism Assessment}

\author{
   Jo\'e Napolitano  \quad Pascal Nguyen\\
  AMIAD\\
  \texttt{napolitanojoe07@gmail.com} \quad
  \texttt{pascal.nguyen@intradef.gouv.fr}
}

\date{}

\begin{document}
\maketitle

\begin{abstract}

Evaluating the realism of generated images remains a fundamental challenge in generative modeling. Existing distributional metrics such as the Fr\'echet Inception Distance (FID)~[1] and CLIP-MMD (CMMD)~[2] compare feature distributions at a semantic level but may overlook fine-grained textural information that can be relevant for distinguishing real from generated images. We introduce \gmmd{} (\gmmdshort{}), a realism metric that leverages Gram matrices computed from intermediate activations of pretrained backbone networks to capture correlations between feature maps. By extracting the upper-triangular part of these symmetric Gram matrices and measuring the Maximum Mean Discrepancy (MMD) between an anchor distribution of real images and an evaluation distribution, \gmmdshort{} produces a representation that encodes textural and structural characteristics at a finer granularity than global embeddings. To select the hyperparameters of the metric, we employ a meta-metric protocol based on controlled degradations applied to MS-COCO~[24] images, measuring monotonicity via Spearman's rank correlation~[4] and Kendall's~$\tau$~[25]. We conduct experiments on both the KADID-10k database~[3] and the RAISE realness assessment dataset~[5] using various backbone architectures, including DINOv2~[6], DC-AE~[7], Stable Diffusion's VAE encoder~[8], VGG19~[9], and the AlexNet backbone from LPIPS~[19], among others. We also demonstrate on a cross-domain driving scenario (KITTI~[26] / Virtual KITTI~[27] / Stanford Cars~[28]) that CMMD can incorrectly rank real images as less realistic than synthetic ones due to its semantic bias, while \gmmdshort{} preserves the correct ordering. Our results suggest that \gmmdshort{} captures complementary information to existing semantic-level metrics.
\end{abstract}

\section{Introduction}
\label{sec:introduction}

The rapid progress of generative image models, from Generative Adversarial Networks (GANs)~[10] to diffusion models~[11, 8], has made it increasingly difficult both to distinguish synthetic images from real photographs and to rank their relative degree of realism. As these models find
applications in content creation, data augmentation, and scientific visualization, the need for reliable automated metrics to quantify image realism has become critical. Human evaluation, while often considered the gold standard, is inherently expensive, slow, and difficult to scale to the large volumes of images produced by modern generative pipelines~[12], motivating the development of automated alternatives.

The most widely adopted automated metric, the Fr\'echet Inception Distance (FID)~[1], estimates the distance between two distributions of Inception-v3 features by assuming both follow multivariate Gaussian distributions and computing the Fr\'echet distance between them. Despite its popularity, FID suffers from well-documented limitations: the Gaussian assumption is often violated in practice, the metric exhibits poor sample efficiency, and the Inception-v3 embeddings (trained on only one million ImageNet images across one thousand classes) fail to capture the rich and diverse content generated by modern text-to-image models~[2]. To address these issues, Jayasumana et al.~[2] proposed CMMD, which replaces Inception features with CLIP embeddings and the Fr\'echet distance with the Maximum Mean Discrepancy (MMD)~[14]. CMMD makes no distributional assumptions, is an unbiased estimator, and leverages richer CLIP representations trained on 400 million image-text pairs. However, like FID, CMMD relies on global, semantic-level embeddings: CLIP is trained via contrastive learning between images and text descriptions~[15], meaning its representations are optimized for semantic alignment rather than for capturing fine-grained visual properties. As a result, both metrics may be less sensitive to textural patterns (micro-textures, noise characteristics, and local structural regularities) that can be informative for assessing image realism.

This observation motivates our work. The Gram matrix of feature activations, introduced by Gatys et al.~[16] for neural style transfer, captures second-order statistics by computing the correlations between all pairs of feature maps at a given layer. These correlations encode textural information independently of spatial layout, making them a candidate for characterizing image properties at a level of granularity that global embeddings do not directly address.

We propose \gmmd{} (\gmmdshort{}), a distributional metric built on three components: (1)~extraction of intermediate feature activations using a pretrained backbone network, (2)~construction and vectorization of Gram matrices from these activations, and (3)~computation of the MMD between the Gram-vector distributions of an anchor set of real images and an evaluation set. To select the hyperparameters of the metric, we design a meta-metric protocol based on controlled degradations applied to images from the MS-COCO dataset~[24]: for each degradation type and severity level, we measure whether the metric responds monotonically, using Spearman's rank correlation coefficient~[4] and Kendall's~$\tau$~[25]. We conduct experiments on the KADID-10k database~[3], which provides human opinion scores on perceptual quality degradations, and on the RAISE dataset~[5] for realism assessment of AI-generated images. Furthermore, we show on a cross-domain driving experiment (using KITTI~[26] as real anchor, Virtual KITTI~[27] as synthetic evaluation, and Stanford Cars~[28] as real evaluation) that CMMD can invert the expected ordering between real and synthetic distributions due to its reliance on semantic embeddings, while \gmmdshort{} correctly identifies the synthetic distribution as more distant from the anchor. We explore various backbone architectures, including a self-supervised Vision Transformer (DINOv2~[6]), high-compression autoencoders (DC-AE~[7]), the VAE encoder from Stable Diffusion~[8], the AlexNet backbone from LPIPS~[19], as well as the classical CNN-based feature extractor VGG19~[9], among others.

The remainder of this paper is organized as follows. Section~\ref{sec:related} reviews related work on image quality metrics and distributional distance measures. Section~\ref{sec:method} details the \gmmdshort{} pipeline, from feature extraction through Gram matrix construction to MMD computation, and describes our meta-metric evaluation framework. Section~\ref{sec:experiments} presents experimental results across backbones and degradation types, and compares \gmmdshort{} with CMMD. Finally, Section~\ref{sec:conclusion} summarizes our findings.

\section{Related Work}
\label{sec:related}
\subsection{Feature Representations for Image Evaluation}
\label{sec:rw_metrics}

Most distributional metrics for evaluating generative models (IS~[12], FID~[1], KID~[18], CMMD~[2]) extract features from the final or near-final layers of pretrained networks, where representations are known to encode high-level semantic content. Visualization studies by Zeiler and Fergus~[23] have shown that CNN layers exhibit a clear hierarchy: early layers respond to edges, colors, and textures, intermediate layers capture local patterns and textures, while deeper layers encode object-level and semantic information. This suggests that restricting feature extraction to the last layers may discard fine-grained textural information relevant to realism assessment.

A natural question is therefore whether using intermediate layers, or altogether different types of encoders, could yield richer representations for this task. The work of Zhang et al.~[19] on LPIPS demonstrated that intermediate activations of classification networks (specifically AlexNet, linearly calibrated on human perceptual judgments) provide multi-layer features that correlate strongly with human perception, illustrating the value of intermediate representations for perceptual quality. We exploit the resulting perceptually-calibrated AlexNet backbone as one of our feature extractors, repurposing it in a distributional rather than pairwise setting.

Beyond CNNs trained on classification tasks, other backbone architectures offer different inductive biases. Variational Autoencoder (VAE) encoders, such as those used in latent diffusion models~[8] or DC-AE~[7], are trained to reconstruct images from a compressed latent space. Their intermediate representations must therefore preserve not only semantic but also low-level structural information necessary for faithful reconstruction. Self-supervised Vision Transformers like DINOv2~[6] learn general-purpose visual features without any text supervision, providing representations that are not biased toward language-aligned semantics. In this work, we systematically compare features from these diverse backbone families (CNNs, VAE encoders, and self-supervised transformers) to assess which intermediate representations best support texture-aware realism metrics.

\subsection{Gram Matrices in Deep Learning}
\label{sec:rw_gram}

The use of Gram matrices to characterize image style was introduced by Gatys et al.~[16] in the context of neural style transfer. Given the activation tensor at a convolutional layer with $C$ channels and $H \times W$ spatial dimensions, the Gram matrix $G \in \reals^{C \times C}$ is computed as the inner product between all pairs of vectorized feature maps, capturing second-order correlations independently of spatial layout. This representation has proven effective for encoding texture~[16] and has since been employed in perceptual loss functions for image synthesis and super-resolution.

While most prior applications of Gram matrices focus on style matching between individual image pairs, we propose to use them as distributional descriptors: each image in a set yields a Gram-derived vector, and the distance between sets is measured via MMD.

\subsection{Maximum Mean Discrepancy}
\label{sec:rw_mmd}

The Maximum Mean Discrepancy (MMD)~[14] is a kernel-based distance between probability distributions that makes no parametric assumptions. Given a characteristic kernel $k$, the squared MMD between two distributions $P$ and $Q$ is defined as:
\begin{multline}
\label{eq:mmd}
\text{MMD}^2(P, Q) = \underset{x,x' \sim P}{\mathbb{E}}\!\big[k(x,x')\big] + \underset{y,y' \sim Q}{\mathbb{E}}\!\big[k(y,y')\big] \\
 - \; 2\,\underset{\substack{x \sim P,\, y \sim Q}}{\mathbb{E}}\!\big[k(x,y)\big].
\end{multline}
Common kernel choices include the Gaussian RBF kernel and the polynomial kernel:
\begin{align}
\label{eq:rbf}
k_{\text{RBF}}(x, y) &= \exp\!\left(-\frac{\|x - y\|^2}{2\sigma^2}\right), \\[4pt]
\label{eq:poly}
k_{\text{poly}}(x, y) &= \left(\tfrac{1}{d}\, x^\top y + 1\right)^3.
\end{align}
With a characteristic kernel such as the Gaussian RBF, MMD equals zero if and only if $P = Q$~[14]. In the image generation literature, KID~[18] uses $k_{\text{poly}}$ on Inception features, while CMMD~[2] uses $k_{\text{RBF}}$ on CLIP embeddings.

A practical challenge arises from the high dimensionality of vectorized Gram matrices: for a backbone with $d$ channels, the resulting vector has dimension $d(d{+}1)/2$, which can reach tens of thousands. Dimensionality reduction methods such as PCA, t-SNE, and UMAP yielded poor results in our experiments and were therefore not retained.

\subsection{Meta-metrics and Evaluation Protocols}
\label{sec:rw_meta}

Evaluating an evaluation metric (a meta-metric problem) requires a principled framework. A natural desideratum is that a realism metric should respond monotonically to controlled degradations: as image quality decreases, the metric should consistently increase. This monotonicity is typically quantified using Spearman's rank correlation coefficient (SROCC)~[4], the standard measure in image quality assessment~[22]. Kendall's~$\tau$~[25] provides a complementary assessment of pairwise ordering consistency.

Our evaluation proceeds in three stages. First, to select the hyperparameters of \gmmdshort{} (backbone, layer, bandwidth), we design a dedicated meta-metric protocol using images from the MS-COCO dataset~[24], to which we apply controlled degradations at multiple severity levels. We then conduct experiments on the KADID-10k database~[3], which contains approximately 10,000 degraded images with subjective quality ratings, enabling comparison with other metrics such as CMMD, and on the RAISE dataset~[5], which pairs real and AI-generated images with subjective realness scores collected through psychovisual studies.

\vspace{2cm}


\section{Methodology}
\label{sec:method}
\begin{figure*}[t]
\centering
\IfFileExists{pipelineGrammmd.png}{\includegraphics[width=1\textwidth]{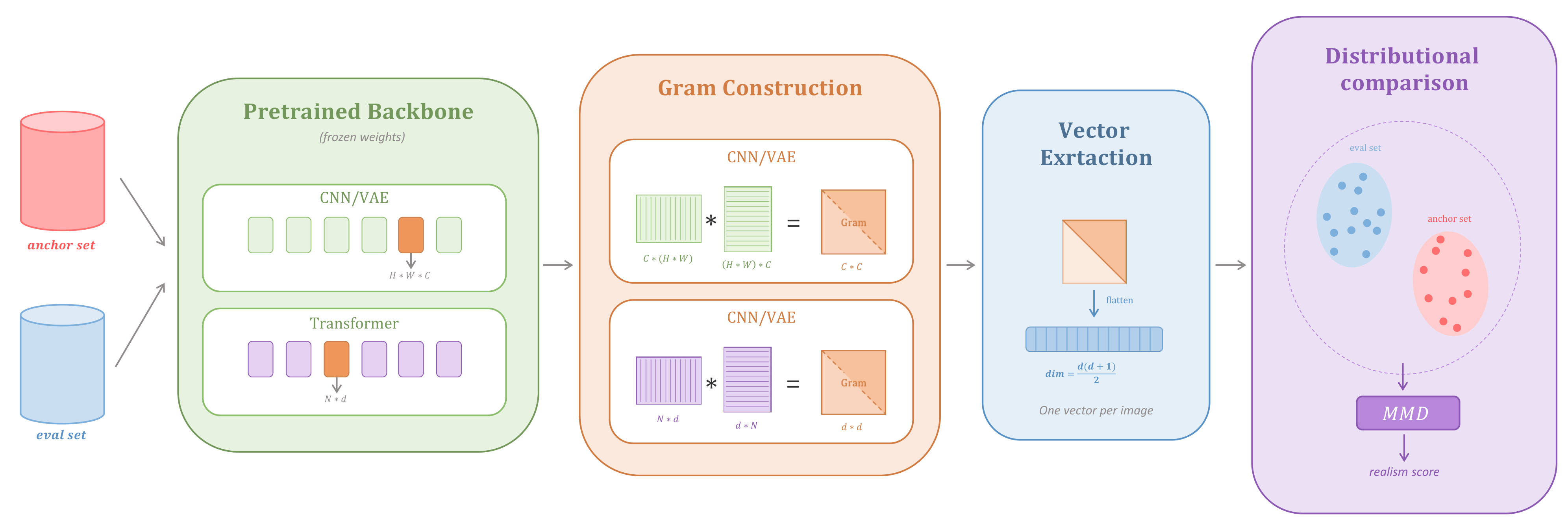}}{\placeholder{6cm}{pipelineGrammmd.png}}
\caption{Overview of the \gmmdshort{} pipeline. Images from the anchor and evaluation sets are passed through a pretrained backbone. Per-pixel (CNN) or per-patch (Transformer) Gram matrices vectorized via the upper-triangular part, and compared using MMD.}
\label{fig:pipeline}
\end{figure*}

This section describes the \gmmdshort{} pipeline in detail. An overview is provided in Figure~\ref{fig:pipeline}.

\subsection{Feature Extraction from Intermediate Layers}
\label{sec:method_features}

Given an input image $I$, we extract the activation tensor at an intermediate layer $l$ of a pretrained backbone network. Depending on the architecture, this tensor takes one of two forms:
\begin{itemize}[nosep,leftmargin=*]
\item \textbf{CNN-based backbones} (VGG19, ResNet, DC-AE, SD-VAE, etc.): the activation at layer $l$ is a tensor $\mathbf{A} \in \reals^{H \times W \times d}$, where $H \times W$ is the spatial resolution and $d$ is the number of channels. This includes VAE encoders such as DC-AE~[7] and the Stable Diffusion VAE~[8], whose convolutional architectures produce spatial feature maps.
\item \textbf{Transformer backbones} (DINOv2, etc.): the activation consists of $N$ patch tokens, each of dimension $d$, yielding a matrix $\mathbf{A} \in \reals^{N \times d}$.
\end{itemize}
The choice of layer $l$ is a key parameter of the metric, which we select empirically as described in Section~\ref{sec:method_meta}. As discussed in Section~\ref{sec:rw_metrics}, different layers generally encode different types of information: in CNN-based architectures, early layers tend to capture low-level textures and edges, while deeper layers encode more semantic content~[23]. For Vision Transformers, Raghu et al.~[29] showed that representations are more uniform across layers than in CNNs, with global information already present in shallow layers; the optimal layer is therefore determined empirically for each backbone.

\subsection{Gram Matrix Construction}
\label{sec:method_gram}

For each spatial position (CNN) or patch (Transformer), we compute a local Gram matrix that captures the pairwise correlations between feature channels.

\paragraph{CNN case.} Let $\mathbf{F} \in \reals^{d \times (H \cdot W)}$ denote the matrix whose columns are the $d$-dimensional feature vectors at each spatial position. The image-level Gram matrix is:
\begin{equation}
\label{eq:gram_cnn}
\bar{G} = \frac{1}{H \cdot W}\, \mathbf{F}\, \mathbf{F}^\top \;\in\; \reals^{d \times d}.
\end{equation}

\paragraph{Transformer case.} Let $\mathbf{P} \in \reals^{d \times N}$ denote the matrix whose columns are the $N$ patch token vectors. The image-level Gram matrix is:
\begin{equation}
\label{eq:gram_tf}
\bar{G} = \frac{1}{N}\, \mathbf{P}\, \mathbf{P}^\top \;\in\; \reals^{d \times d}.
\end{equation}

In both cases, the matrix product averages over spatial positions or patches, removing spatial layout information and retaining only the statistical correlations between feature channels. This is by design: the resulting matrix $\bar{G}$ captures \emph{which} textural patterns co-occur in the image, independently of \emph{where} they appear. This removal of spatial information prevents images with similar spatial arrangements from being considered overly similar in the embedding space, ensuring the metric focuses on texture rather than layout.

\subsection{Vectorization}
\label{sec:method_vec}

Since $\bar{G}$ is symmetric ($\bar{G} = \bar{G}^\top$), it contains redundant information below the diagonal. We extract the upper-triangular part (including the diagonal) and flatten it into a vector:
\begin{equation}
\label{eq:vec}
\mathbf{v} = \mathrm{upper\text{-}tri}(\bar{G}) \;\in\; \reals^{d(d+1)/2}.
\end{equation}
Each image in the dataset is thus represented by a single vector $\mathbf{v}$. A key property of the Gram matrix representation is that it produces embeddings of identical dimensionality for all images, regardless of their spatial resolution, since $\bar{G} \in \reals^{d \times d}$ depends only on the number of channels $d$. This allows the application of MMD in a shared embedding space across images of different sizes.

Given an anchor set of $N_A$ images and an evaluation set of $N_E$ images, we obtain two collections of vectors: $\{\mathbf{v}^{(a)}_i\}_{i=1}^{N_A}$ and $\{\mathbf{v}^{(e)}_j\}_{j=1}^{N_E}$.

\paragraph{Standardization.} Gram-vector components can span several orders of magnitude, causing the RBF kernel to collapse to zero for a fixed $\gamma$. We therefore standardize all vectors using the mean and standard deviation estimated from the anchor set:
\begin{equation}
\label{eq:standardize}
\tilde{\mathbf{v}} = \frac{\mathbf{v} - \boldsymbol{\mu}}{\boldsymbol{\sigma}},
\end{equation}
which brings pairwise distances to a consistent scale without altering their rank ordering.

\subsection{MMD Computation}
\label{sec:method_mmd}

The \gmmdshort{} score is the squared MMD between the anchor and evaluation distributions in the Gram-vector space. Using the Gaussian RBF kernel $k$ defined in~(\ref{eq:rbf}), the unbiased empirical estimator is:
\begin{align}
\label{eq:mmd_empirical}
\widehat{\text{MMD}}^2 &= \frac{1}{N_A(N_A{-}1)} \!\sum_{i \neq i'} k\!\left(\mathbf{v}^{(a)}_i, \mathbf{v}^{(a)}_{i'}\right) \notag\\
&+ \frac{1}{N_E(N_E{-}1)} \!\sum_{j \neq j'} k\!\left(\mathbf{v}^{(e)}_j, \mathbf{v}^{(e)}_{j'}\right) \notag\\
&- \frac{2}{N_A N_E} \sum_{i,j} k\!\left(\mathbf{v}^{(a)}_i, \mathbf{v}^{(e)}_j\right).
\end{align}

\paragraph{Bandwidth parameter.} The bandwidth $\sigma$ of the RBF kernel controls the scale at which distributional differences are captured. A common heuristic sets $\sigma$ to the median of pairwise distances within the anchor set~[14]. In the remainder of this paper, we use the parametrization $\gamma = 1/(2\sigma^2)$ and define the \emph{median heuristic} value as:
\begin{equation}
\label{eq:gamma_median}
\gamma_{\mathrm{med}} = \frac{1}{2\,\mathrm{median}_{i < j}\!\left(\|\mathbf{v}^{(a)}_i - \mathbf{v}^{(a)}_j\|^2\right)}.
\end{equation}

\subsection{Parameter Selection via Meta-metric Protocol}
\label{sec:method_meta}

The hyperparameters of \gmmdshort{} (backbone and intermediate layer $l$) are selected jointly using a dedicated protocol based on controlled degradations of MS-COCO images. We evaluate two anchor configurations: a \emph{reference anchor}, where the 50 anchor images are the same clean references used to generate the degradations, and an \emph{independent anchor}, where the 50 anchor images are drawn from a disjoint set of MS-COCO images. The procedure is as follows:
\begin{enumerate}[nosep,leftmargin=*]
\item \textbf{Reference images}: 50 MS-COCO~[24] images serving as clean references.
\item \textbf{Degraded images}: each reference image is degraded using 20 distortion types at 10 increasing severity levels (see Appendix for the full list), yielding 200 groups of 50 degraded images.
\item \textbf{Distributional \gmmdshort{}}: for each degradation type $t$ and severity level $s$, the set of 50 degraded images at level $s$ forms an evaluation distribution $\mathcal{I}_{t,s}$. We compute $d_{t,s} = \gmmdshort{}^2(\mathcal{I}_{t,s}, X)$ between this set and the anchor $X$ using~(\ref{eq:mmd_empirical}). For a well-behaved metric, we expect:
\begin{equation}
\label{eq:ordering}
d_{t,1} < d_{t,2} < \cdots < d_{t,10}.
\end{equation}
\item \textbf{Monotonicity evaluation}: for each degradation type $t$, we assess how well the 10 scores respect this ordering using Spearman's $\rho$~[4], which measures global monotonic agreement, and Kendall's $\tau$~[25], which captures whether the pairwise ordering between severity levels is respected. We report the average $\rho$ and $\tau$ across all 20 degradation types.
\item \textbf{Parameter selection}: we select the backbone, layer, and $\gamma$ that jointly maximise the average Spearman's $\rho$ and Kendall's $\tau$.
\end{enumerate}

\paragraph{Search space.} We apply this protocol to all intermediate layers of five backbone families:
SD-VAE (17 layers), DC-AE (16 layers), DINOv2-ViT-B/14 (14 layers),
VGG19 (18 layers), and LPIPS-VGG (13 layers).
For each layer, we evaluate 10 values of $\gamma$: the median heuristic
$\gamma_{\mathrm{med}}$~\cite{gretton2012kernel} and nine multiples thereof
($0.01\times$, $0.03\times$, $0.1\times$, $0.3\times$, $0.5\times$,
$2\times$, $5\times$, $10\times$, $30\times\,\gamma_{\mathrm{med}}$),
yielding 780 configurations in total.

\subsection{Meta-metric Results and Selected Configurations}
\label{sec:method_results}

\begin{figure*}[htbp]
\centering
\begin{minipage}[t]{0.48\textwidth}
\centering
\IfFileExists{figure3rhoprofiles.png}{\includegraphics[width=\textwidth]{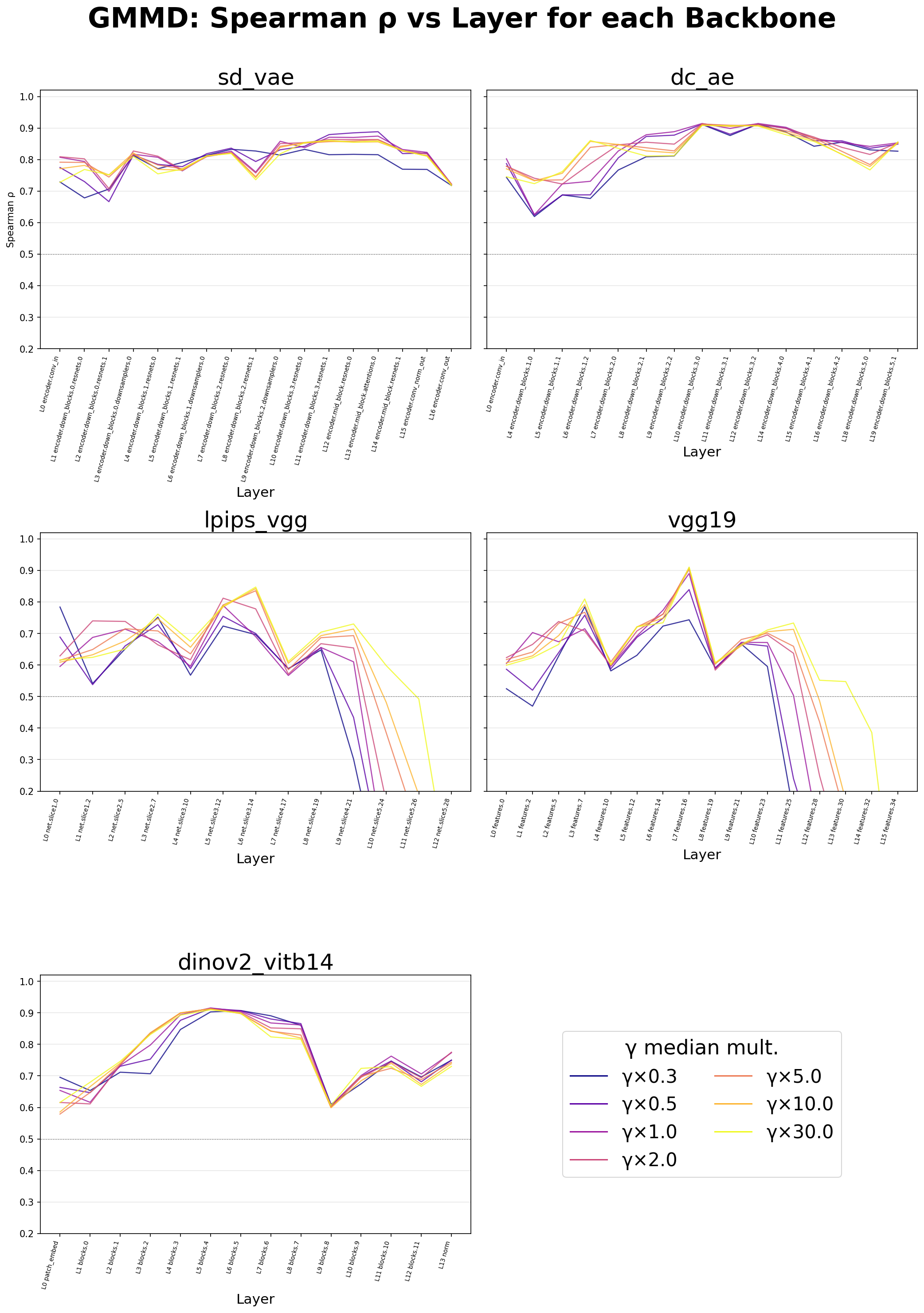}}{\placeholder{5cm}{figure3rhoprofiles.png}}
\vspace{-0.5em}
\caption{Spearman $\rho$ vs.\ layer index for each backbone, top-7 $\gamma$ values (decreasing opacity) and best configuration~($\bigstar$).}
\label{fig:layer_profiles}
\end{minipage}%
\hfill
\begin{minipage}[t]{0.48\textwidth}
\centering
\IfFileExists{figure2rhogamma.png}{\includegraphics[width=\textwidth]{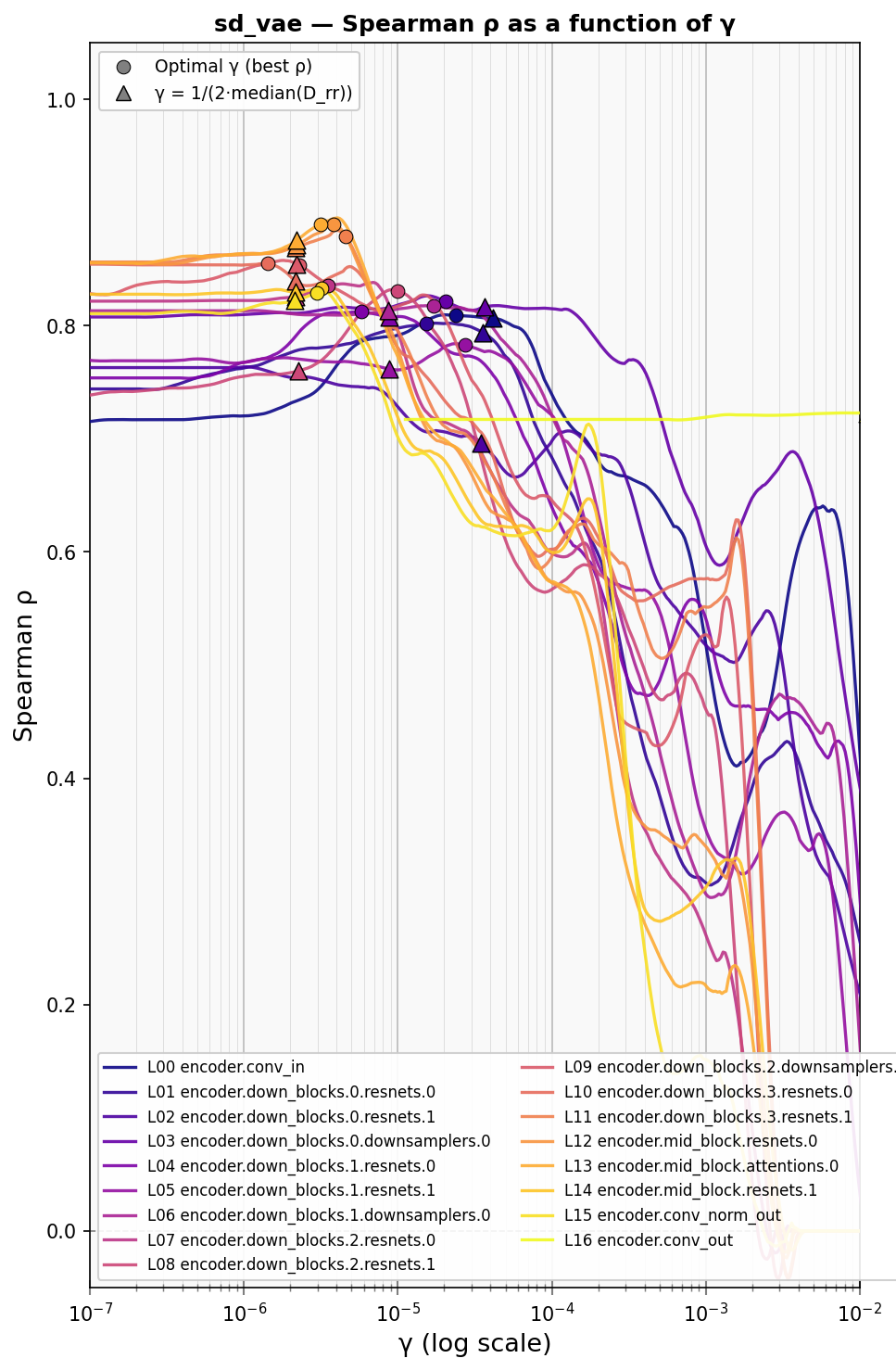}}{\placeholder{5cm}{figure2rhogamma.png}}
\vspace{-0.5em}
\caption{Spearman $\rho$ vs.\ $\gamma$ for the 17 SD-VAE layers. Triangles: $\gamma_{\mathrm{med}}$; circles: empirical optimum.}
\label{fig:gamma_srocc}
\end{minipage}
\end{figure*}

\paragraph{Layer profiles.}
Figure~\ref{fig:layer_profiles} plots $\rho$ against layer index for each backbone, showing the top-7 $\gamma$ values with decreasing opacity. For the two VAE encoders (SD-VAE and DC-AE), performance improves progressively with depth, with the $\gamma$ curves largely overlapping, indicating robustness to bandwidth choice once the layer is well selected. VGG19 and LPIPS-VGG are less consistent: performance depends strongly on both the layer and $\gamma$, though VGG19 tends to perform better at earlier layers. DINOv2, despite higher variance across $\gamma$ values (reflecting the more uniform depth hierarchy of Vision Transformers~[29]) achieves competitive peak performance.

\paragraph{Bandwidth analysis.}
Figure~\ref{fig:gamma_srocc} plots Spearman's $\rho$ against $\gamma$ for each of the 17 SD-VAE layers. Triangles mark the median heuristic $\gamma_{\mathrm{med}}$; circles mark the empirically optimal $\gamma$. The median heuristic is consistently positioned slightly to the left of the optimum, which typically lies between $\gamma_{\mathrm{med}}$ and $10\times\gamma_{\mathrm{med}}$. Beyond this range, the kernel becomes too narrow: it effectively only matches quasi-identical points in feature space, losing sensitivity to global texture and style information. This collapse regime is generally observed past $10\times\gamma_{\mathrm{med}}$.

\begin{figure*}[htbp]
\centering
\begin{minipage}[t]{0.48\textwidth}
\centering
\IfFileExists{figure4top20_spearman_kendall.png}{\includegraphics[width=\textwidth]{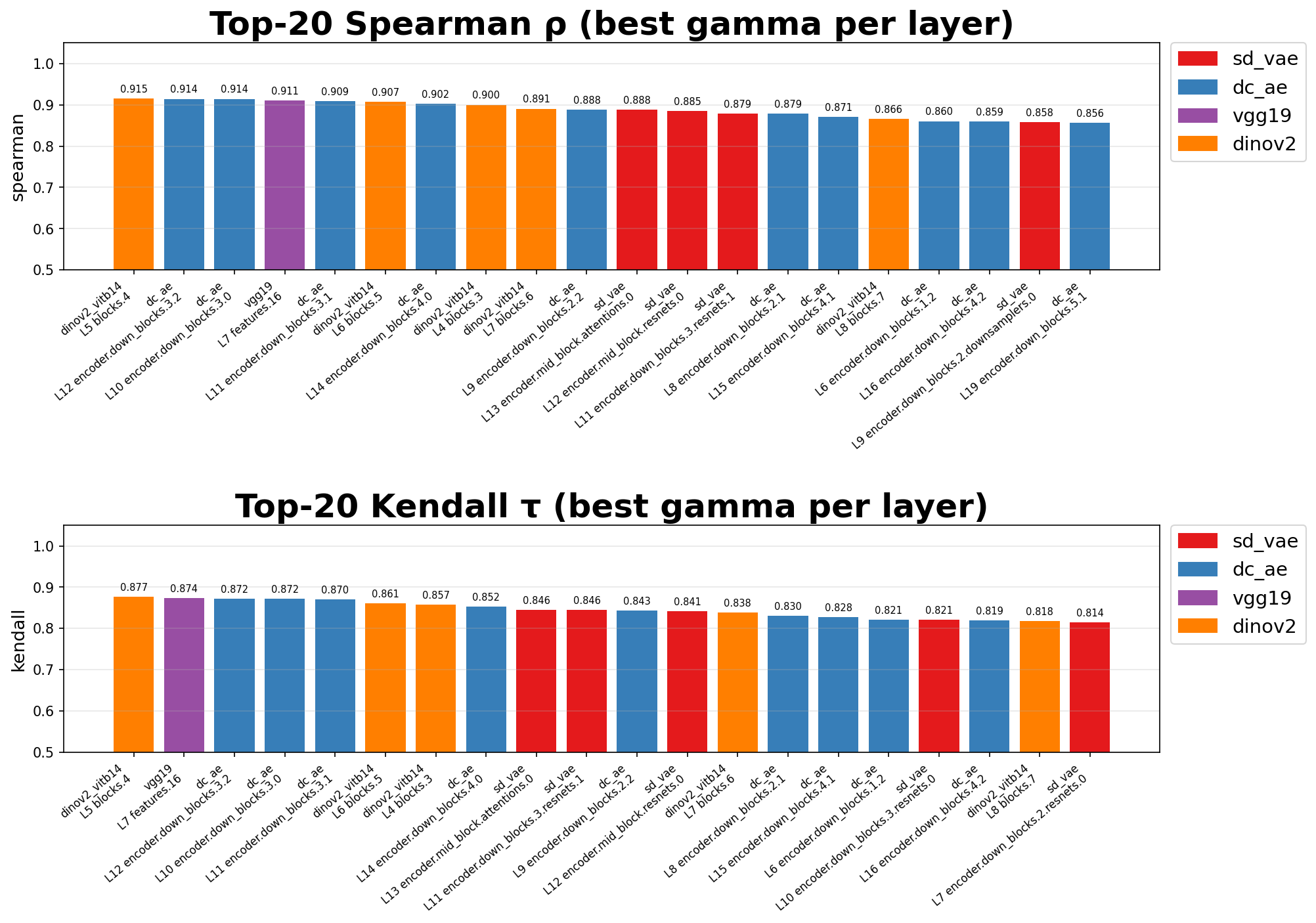}}{\placeholder{5cm}{figure4top20\_spearman\_kendall.png}}
\vspace{-0.5em}
\caption{Top-20 configurations ranked by Spearman's $\rho$ (left) and Kendall's $\tau$ (right), coloured by backbone.}
\label{fig:top20}
\end{minipage}%
\hfill
\begin{minipage}[t]{0.48\textwidth}
\centering
\IfFileExists{figure5_boxplot_per_backbone.png}{\includegraphics[width=\textwidth]{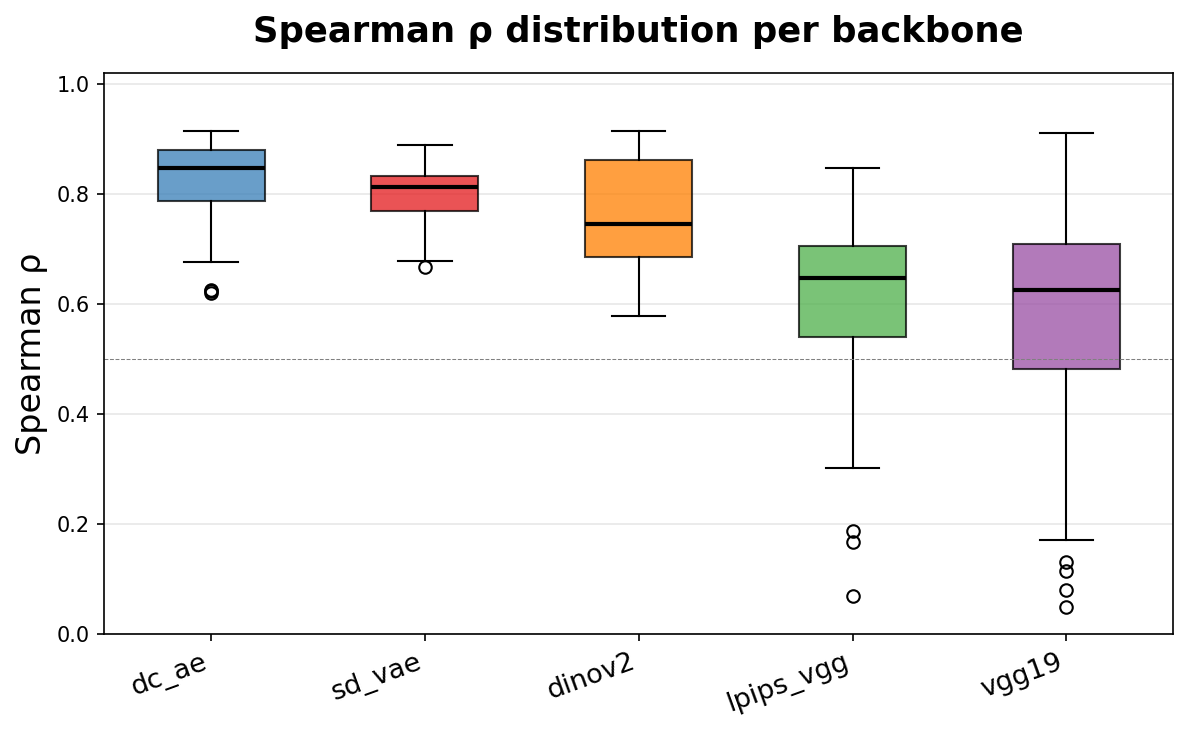}}{\placeholder{5cm}{figure5\_boxplot\_per\_backbone.png}}
\vspace{-0.5em}
\caption{Distribution of Spearman's $\rho$ across all configurations, grouped by backbone.}
\label{fig:boxplot}
\end{minipage}
\end{figure*}

\paragraph{Top configurations.}
Figure~\ref{fig:top20} ranks the top-20 configurations by Spearman's $\rho$ and by Kendall's $\tau$. DC-AE accounts for 11 of the 20 best configurations, while SD-VAE contributes 4 --5 depending on the criterion-- confirming the strong and stable performance of VAE encoders. DINOv2, although more dispersed across layers (Figure~\ref{fig:boxplot}), achieves good results between layers L4 and L8. VGG19 appears only once in the top-20 and LPIPS not at all, indicating that these backbones lack the consistency required for reliable deployment.

\paragraph{Selected configurations.}
Based on this analysis, we retain three reference configurations for the experiments in Section~\ref{sec:experiments}: DINOv2 L5, DC-AE L12, and SD-VAE L13. In each case, $\gamma$ is set in the range between $0.1 \times \gamma_{\mathrm{med}}$ and $10 \times \gamma_{\mathrm{med}}$, safely below the collapse threshold.

\section{Experiments}
\label{sec:experiments}

We compare \gmmdshort{} with CMMD~[2] across three experiments. First, on KADID-10k~[3], where human quality ratings provide a perceptual ground truth for image degradations. Second, on RAISE~[5], where subjective realness scores allow evaluation on AI-generated images. Third, on a cross-domain driving scenario using KITTI~[26], Virtual KITTI~2 [27], and Stanford Cars~[28], where we show that CMMD can confuse synthetic and real distributions due to its semantic bias.

\subsection{Evaluation on KADID-10k}
\label{sec:exp_kadid}

KADID-10k~[3] comprises 81 pristine reference images, each degraded by 25 distortion types with 5 severity levels, yielding 10{,}125 distorted images with associated DMOS (Differential Mean Opinion Score). Higher DMOS corresponds to stronger perceived degradation. We reorganise the 10{,}125 images into 125 groups of 81 images, ranked by DMOS: group~0 contains the most severely degraded images (lowest DMOS), group~124 the least degraded (highest DMOS). For the anchor, we use 1{,}000 natural images randomly sampled from MS-COCO~[24]. For each group of rank $k \in \{0,\ldots,124\}$, we compute $\gmmdshort{}^2$ between the group distribution and the anchor. We evaluate the correlation between group rank and the metric score using Spearman's $\rho$ and Kendall's $\tau$, and perform the same evaluation with CMMD (using $\gamma = 0.005$ as prescribed by~[2]).

\begin{figure}[htbp]
\centering
\IfFileExists{figure6scatter.png}{\includegraphics[width=\columnwidth]{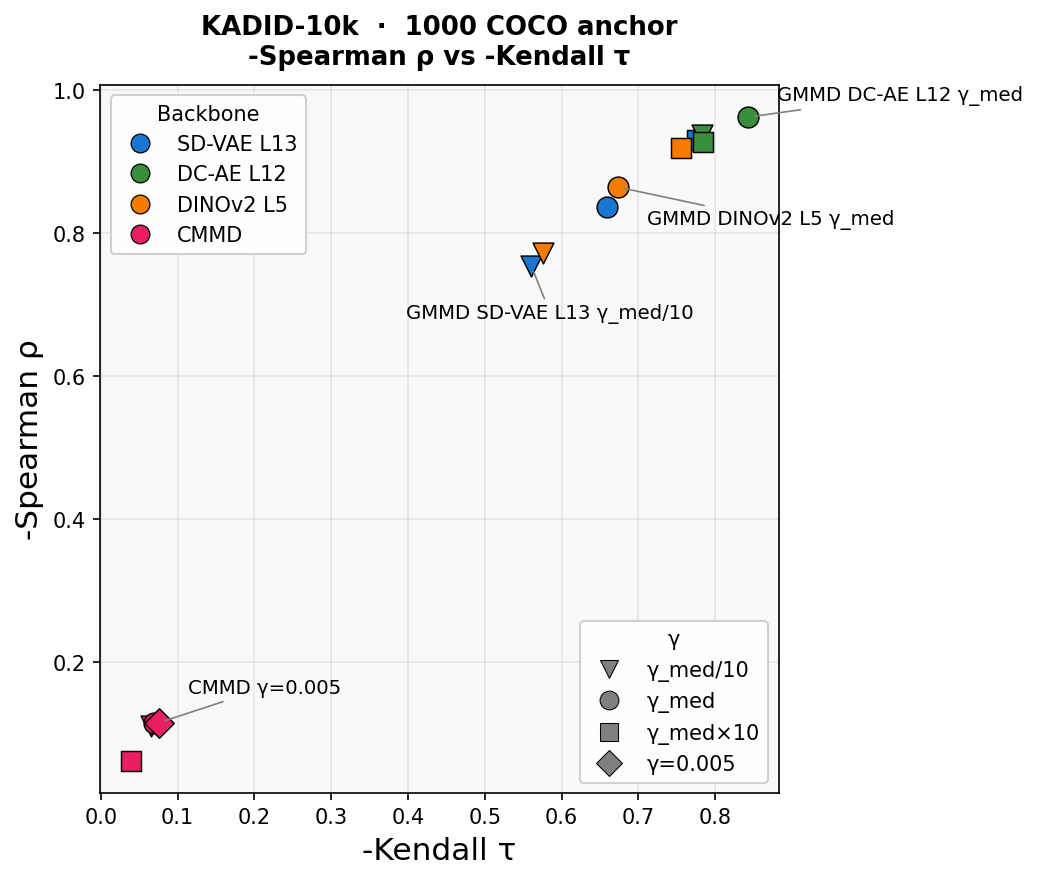}}{\placeholder{5cm}{figure6scatter.png}}
\caption{Spearman's $\rho$ vs.\ Kendall's $\tau$ on KADID-10k for each configuration.}
\label{fig:scatter_kadid}
\end{figure}

\begin{figure}[htbp]
\centering
\IfFileExists{figure7comibnedselected.png}{\includegraphics[width=\columnwidth]{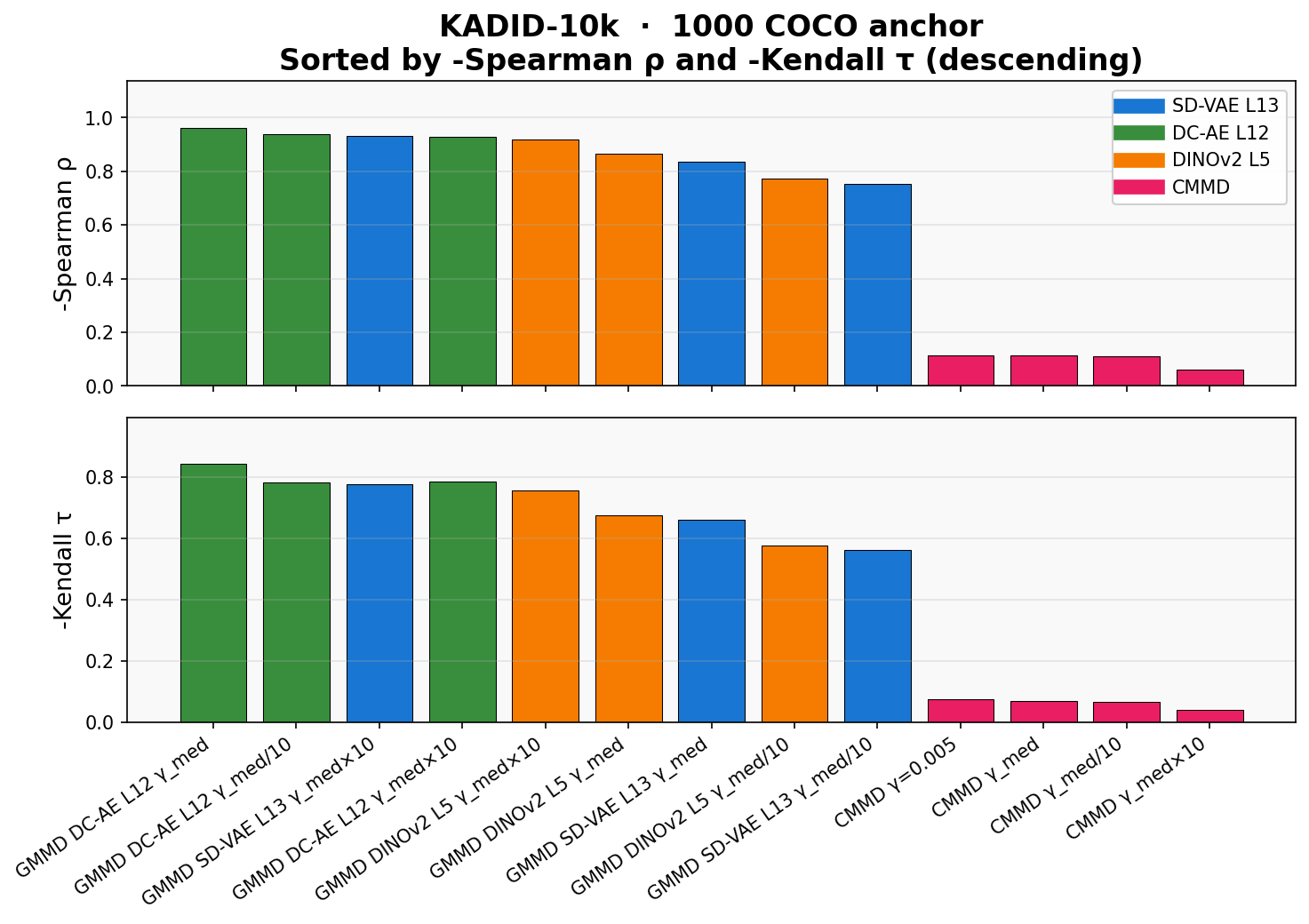}}{\placeholder{5cm}{figure7comibnedselected.png}}
\caption{Spearman's $\rho$ (top) and Kendall's $\tau$ (bottom) on KADID-10k across our 13 configurations and CMMD~[2].}
\label{fig:bar_kadid}
\end{figure}

Figures~\ref{fig:scatter_kadid} and~\ref{fig:bar_kadid} show that \gmmdshort{} consistently outperforms CMMD on both Spearman's $\rho$ and Kendall's $\tau$. Both correlation coefficients are negative, as expected: as DMOS increases (stronger degradation), the MMD (a positive distance) decreases, since more degraded images move further from the real anchor distribution. All three \gmmdshort{} configurations capture this negative correlation more strongly than CMMD across both criteria.

\subsection{Realism Assessment on RAISE}
\label{sec:exp_raise}

RAISE~[5] contains 480 AI-generated images and 120 real images, each paired with a Mean Opinion Score (MOS) collected via a psychovisual study following ITU-T P.910~[22]. Ratings were gathered on a continuous 0--100 scale from at least 23 participants per image. Participants were instructed to rate perceived realness independently of semantic content, making this benchmark well suited to a texture-grounded metric.

We use 1{,}000 MS-COCO~[24] images as the anchor distribution. The 480 AI-generated images are sorted by ascending MOS and partitioned into 24 non-overlapping groups of 20 images. For each group we compute the mean MOS and the \gmmdshort{} score, then evaluate monotonicity using Spearman's $\rho$ and Kendall's $\tau$.

\begin{figure}[htbp]
\centering
\IfFileExists{figure8groupping.png}{\includegraphics[width=\columnwidth]{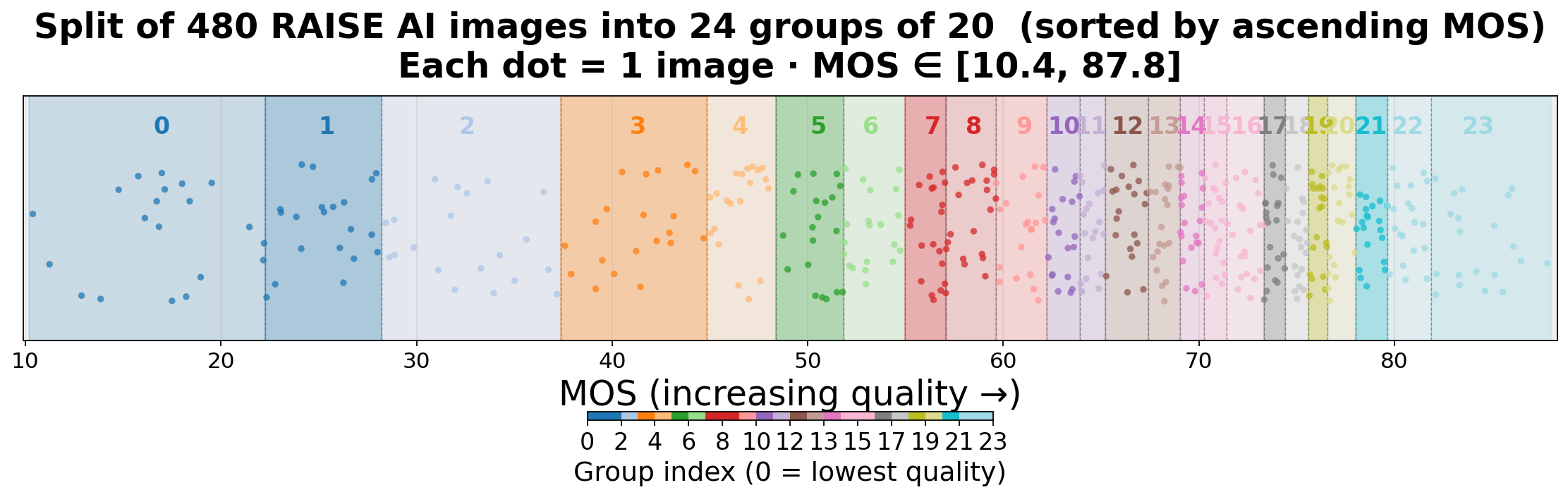}}{\placeholder{5cm}{figure8groupping.png}}
\caption{Partitioning of the 480 AI-generated RAISE images into 24 groups of 20, sorted by ascending MOS.}
\label{fig:raise_groups}
\end{figure}

\begin{figure}[htbp]
\centering
\IfFileExists{figure9regressionraise.png}{\includegraphics[width=\columnwidth]{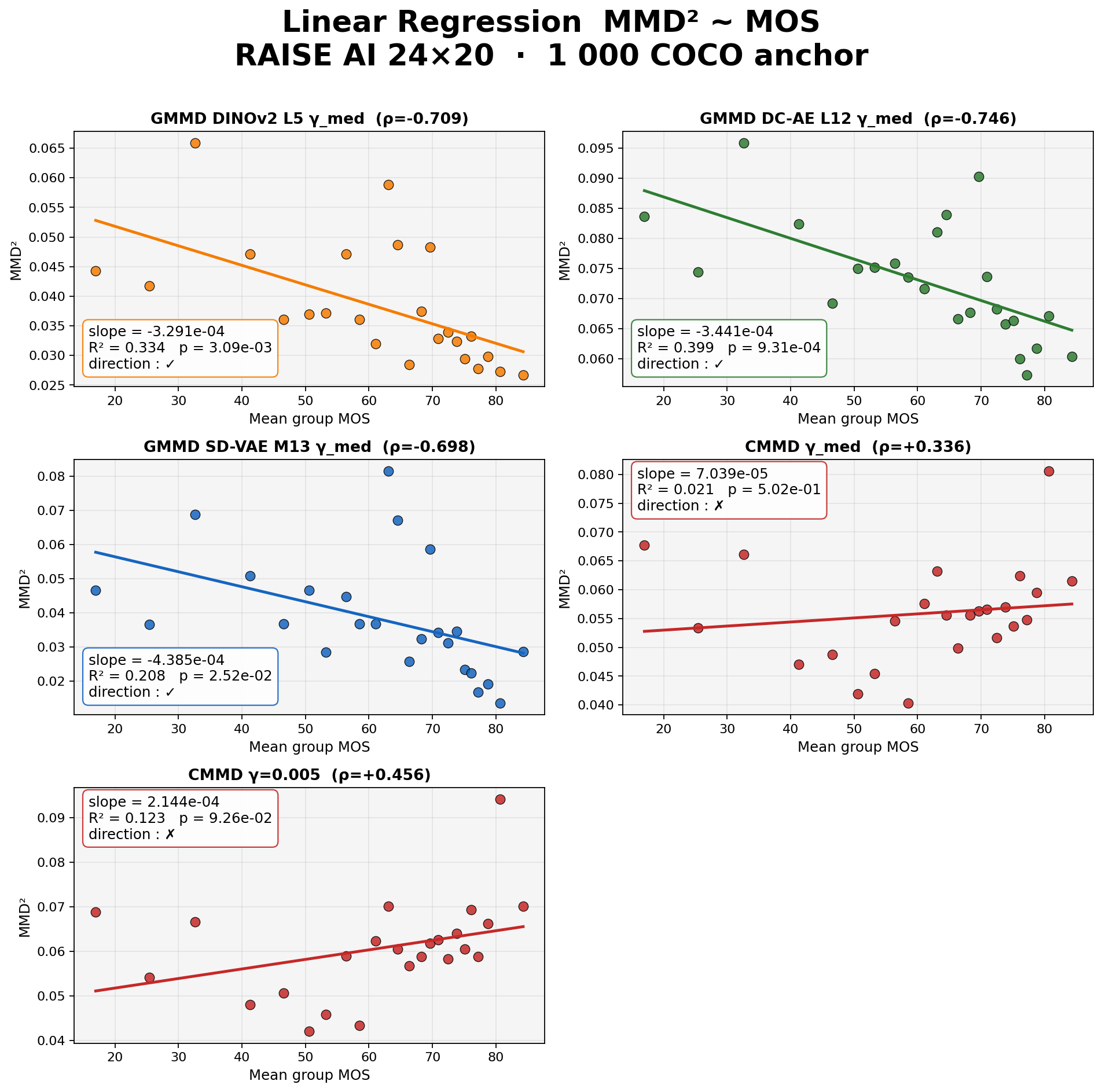}}{\placeholder{5cm}{figure9regressionraise.png}}
\caption{Linear regression between group-averaged MOS and metric score for each \gmmdshort{} configuration and for CMMD (with $\gamma_{\mathrm{med}}$ and $\gamma = 0.005$). Spearman's $\rho$ is reported for each fit.}
\label{fig:raise_regression}
\end{figure}

Figure~\ref{fig:raise_regression} shows that all \gmmdshort{} configurations exhibit a clear negative correlation with MOS: as perceived realism increases, the distributional distance to the real anchor decreases. The Spearman coefficients are negative and statistically significant. For CMMD, the picture is different. At low MOS values (below approximately 60), CMMD discriminates between groups reasonably well. However, for MOS values above 60 (where images approach photorealism), CMMD loses discriminative power and even produces a positive correlation with a non-significant p-value. This is likely caused by the semantic nature of CLIP embeddings: once images are sufficiently realistic, their semantic content dominates their embedding, and CMMD no longer captures the subtle textural differences that separate near-realistic from truly realistic images. \gmmdshort{}, by contrast, places image distributions more accurately along the realism axis.

\subsection{Real vs.\ Synthetic: KITTI Experiment}
\label{sec:exp_kitti}

A fundamental limitation of MMD-based metrics is that they measure distributional \emph{matching}, not whether evaluation samples lie within the support of the anchor distribution. Consider an evaluation set forming a tight cluster entirely contained within the anchor distribution: despite all samples being individually plausible, the MMD remains large because the evaluation set covers only a fraction of the anchor's support. Conversely, a set with broader variability may achieve a lower MMD by better matching the anchor's global spread, even if some of its samples fall outside the anchor's support. MMD is therefore driven by coverage of the anchor distribution, not just inclusion within it. These two scenarios are illustrated in Figure 10. 

\begin{figure}[htbp]
\centering
\includegraphics[width=\columnwidth]{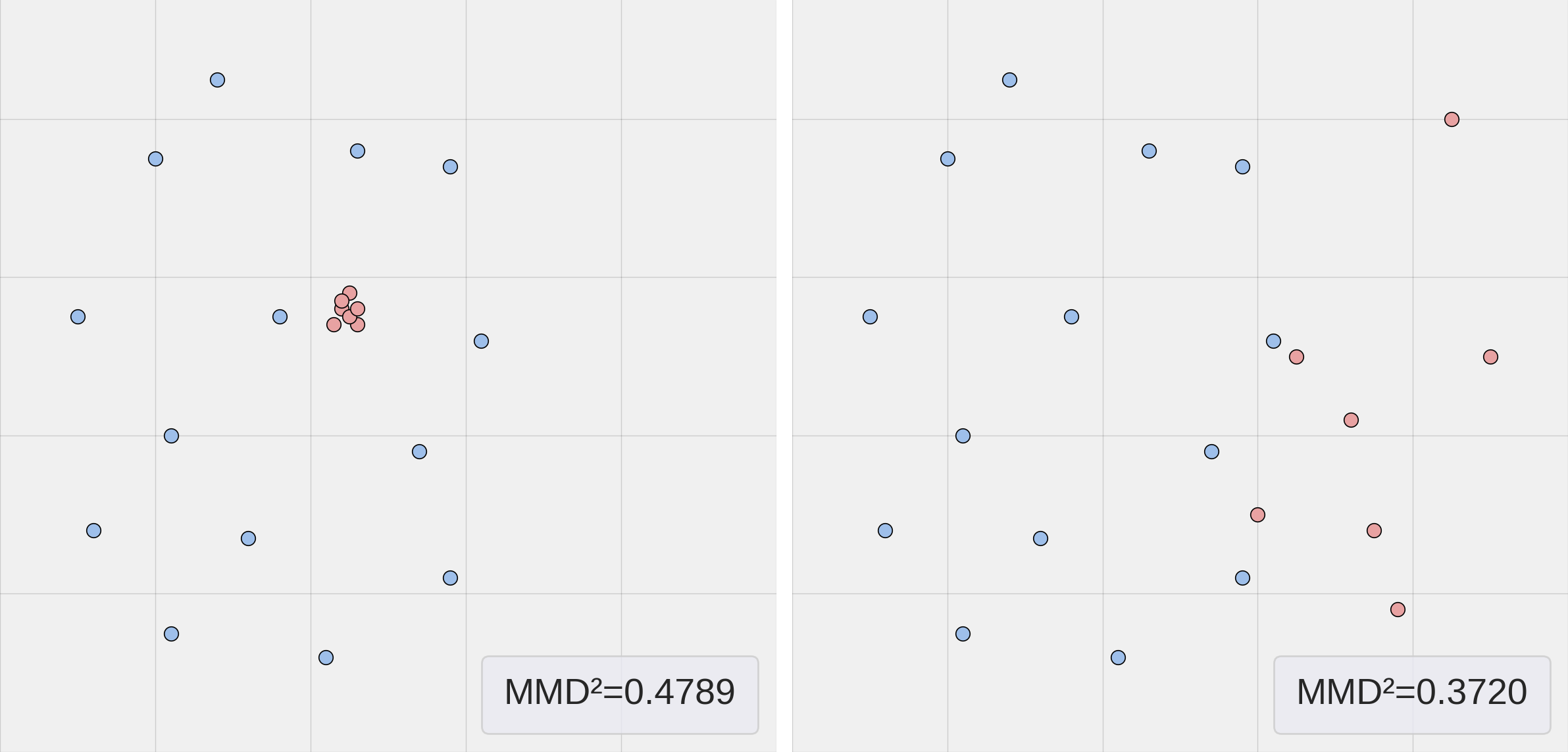}
\caption{Two examples illustrating MMD behaviour. Left: a tight evaluation cluster within the anchor yields a high MMD due to limited coverage. Right: a broader but shifted evaluation set achieves lower MMD by better matching the anchor's spread.}
\label{fig:mmd_illustration}
\end{figure}

We design an experiment to expose this limitation in CMMD. We use KITTI~[26], an autonomous driving dataset of real street-level images, as the anchor distribution (1{,}000 randomly sampled images). We compare two evaluation distributions: (i) Virtual KITTI~2~[27], a synthetic dataset of photo-realistic driving scenes rendered with the Unity engine to replicate KITTI scenarios (1{,}000 images); and (ii) Stanford Cars~[28], a real-world dataset of 16{,}185 photographs of vehicles in diverse settings (1{,}000 randomly sampled images). A realism metric should assign a smaller distance to the real evaluation set (Stanford Cars) than to the synthetic one (Virtual KITTI~2).

\begin{table}[htbp]
\centering
\setlength{\tabcolsep}{6pt}
\caption{\gmmdshort{} (DC-AE L12) and CMMD evaluated on a KITTI anchor (1{,}000 images) against Virtual KITTI~2 (synthetic) and Stanford Cars (real). Ratio~$<1$ indicates inversion.}
\label{tab:inversion}
\resizebox{\columnwidth}{!}{%
\begin{tabular}{|l|c|c|c|c|c|}
\hline
\textbf{Metric} & $\boldsymbol{\gamma}$ & \textbf{VKITTI2} & \textbf{Stanford Cars} & \textbf{Ratio} & \textbf{Inv.?} \\
                &                       & \textit{(synth.)} & \textit{(real)}       &                &                     \\
\hline
\gmmdshort{} (DC-AE L12) & $\gamma_{\mathrm{med}}$ & $2.57 \times 10^{-9}$ & $2.65 \times 10^{-11}$ & $96.7$ & No \\
\hline
CMMD (CLIP ViT-L/14) & $\gamma = 0.005$ & $0.494$ & $0.517$ & $\mathbf{0.96}$ & \textbf{Yes} \\
\hline
\end{tabular}%
}
\end{table}

\begin{figure}[htbp]
\centering
\IfFileExists{figure11grammmddc_aeL12.png}{\includegraphics[width=\columnwidth]{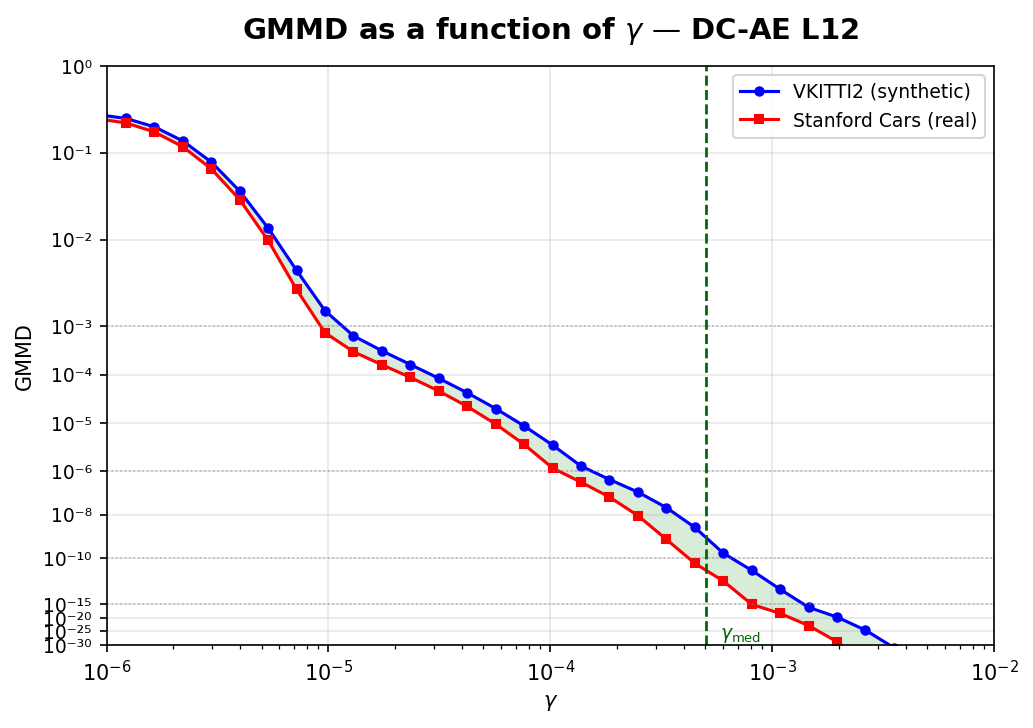}}{\placeholder{5cm}{figure11grammmddc\_aeL12.png}}
\caption{\gmmdshort{} scores for Virtual KITTI~2 (synthetic) and Stanford Cars (real) against a KITTI anchor, across multiple $\gamma$ values. The correct ordering (synthetic $>$ real) is preserved for all $\gamma$.}
\label{fig:kitti_gmmd}
\end{figure}

Table~\ref{tab:inversion} reveals that CMMD assigns a \emph{lower} distance to the synthetic Virtual KITTI~2 distribution than to the real Stanford Cars distribution, revealing an inversion of the expected realism ordering. This occurs because Virtual KITTI~2 was designed to replicate KITTI's driving scenarios and therefore covers a similar semantic space in CLIP's embedding, whereas Stanford Cars, though consisting of real photographs, occupies a narrower region of KITTI's semantic distribution (mostly cars, with little variation in scenes). CMMD penalises this limited semantic coverage as if it were lack of realism.

\gmmdshort{}, by contrast, operates on Gram-based representations that emphasize low-level texture and style. In this space, the real Stanford Cars images exhibit substantial variability in local statistics, producing a well-dispersed distribution that better matches the spread of the KITTI anchor, resulting in a lower distance. As shown in Figure~\ref{fig:kitti_gmmd}, this correct ordering is preserved across all tested $\gamma$ values.

\section{Conclusion}
\label{sec:conclusion}

We introduced \gmmd{} (\gmmdshort{}), a distributional metric for image realism assessment that captures second-order feature correlations through Gram matrices extracted from intermediate layers of pretrained backbones. By vectorizing these matrices and computing MMD in the resulting high-dimensional space, \gmmdshort{} targets textural and structural properties that semantic-level metrics overlook.

Our meta-metric protocol on controlled MS-COCO degradations showed that VAE encoders (SD-VAE, DC-AE) and DINOv2 provide the most reliable backbone representations, with the optimal $\gamma$ consistently located between $\gamma_{\mathrm{med}}$ and $10\times\gamma_{\mathrm{med}}$. On KADID-10k, all \gmmdshort{} configurations outperformed CMMD on both Spearman's $\rho$ and Kendall's $\tau$ for tracking perceptual degradation severity. On RAISE, \gmmdshort{} maintained a significant negative correlation with human realness ratings across the full MOS range, whereas CMMD lost discriminative power for near-realistic images. Finally, on a cross-domain driving experiment, we demonstrated that CMMD can invert the expected ordering between real and synthetic distributions due to its reliance on semantic embeddings, while \gmmdshort{} correctly identified the synthetic distribution as more distant.

These results position \gmmdshort{} as a texture-aware complement to semantic metrics such as CMMD.


\newpage
\onecolumn
\appendix
\section*{Appendix}
\FloatBarrier

\setlength{\textfloatsep}{12pt}   
\setlength{\floatsep}{14pt}       
\setlength{\intextsep}{12pt}      

\noindent\begin{minipage}{\textwidth}
\captionsetup{type=table}
\caption{20 synthetic degradation types $\times$ 10 severity levels applied to the MS-COCO anchor images.
Linear interpolation: level~1~$\to$~minimal parameter, level~10~$\to$~maximal parameter.}
\label{tab:distortions}
\centering
\scriptsize
\setlength{\tabcolsep}{3pt}
\begin{tabular}{clll *{10}{r}}
\toprule
\textbf{\#} & \textbf{KADID} & \textbf{Nom} & \textbf{Param.}
  & \textbf{1} & \textbf{2} & \textbf{3} & \textbf{4} & \textbf{5}
  & \textbf{6} & \textbf{7} & \textbf{8} & \textbf{9} & \textbf{10} \\
\midrule
1  & T1  & Gaussian noise       & $\sigma$
   & .002 & .004 & .006 & .009 & .011 & .013 & .015 & .018 & .020 & .022 \\
2  & T3  & Multiplicative noise & $\sigma$
   & .002 & .005 & .008 & .011 & .014 & .018 & .021 & .024 & .027 & .030 \\
3  & T5  & Brighten             & $\gamma$
   & .960 & .958 & .957 & .955 & .953 & .952 & .950 & .948 & .947 & .945 \\
4  & T6  & Darken               & $\gamma$
   & 1.100 & 1.103 & 1.107 & 1.110 & 1.113 & 1.117 & 1.120 & 1.123 & 1.127 & 1.130 \\
5  & T8  & Jitter               & amp (px)
   & 1.0 & 1.4 & 1.9 & 2.3 & 2.8 & 3.2 & 3.7 & 4.1 & 4.6 & 5.0 \\
6  & T9  & Patches              & $ps / n$
   & 4/1 & 5/2 & 5/2 & 6/3 & 7/3 & 7/4 & 8/4 & 9/5 & 9/5 & 10/6 \\
7  & T10 & Pixelate             & factor
   & 2 & 2 & 2 & 2 & 2 & 3 & 3 & 3 & 3 & 3 \\
8  & T11 & Quantization         & bits
   & 8 & 8 & 7 & 7 & 7 & 6 & 6 & 6 & 5 & 5 \\
9  & T12 & Fog                  & $\alpha$
   & .020 & .030 & .040 & .050 & .060 & .070 & .080 & .090 & .100 & .110 \\
10 & T13 & Color cast cool      & $s$
   & .020 & .027 & .033 & .040 & .047 & .053 & .060 & .067 & .073 & .080 \\
11 & T15 & Chromatic aberr.     & shift (px)
   & 1 & 1 & 1 & 2 & 2 & 2 & 2 & 3 & 3 & 3 \\
12 & T16 & Sparse sampling      & frac
   & .010 & .018 & .026 & .033 & .041 & .049 & .057 & .064 & .072 & .080 \\
13 & T17 & JPEG compress.       & qualit\'e
   & 95 & 92 & 90 & 87 & 85 & 82 & 80 & 77 & 75 & 72 \\
14 & T18 & Gaussian blur        & $\sigma$
   & .200 & .222 & .244 & .267 & .289 & .311 & .333 & .356 & .378 & .400 \\
15 & T19 & Lens blur (disk)     & $r$ (px)
   & 1 & 1 & 1 & 1 & 1 & 2 & 2 & 2 & 2 & 2 \\
16 & T20 & Motion blur (horiz.) & $L$ (px)
   & 3 & 3 & 3 & 3 & 3 & 4 & 4 & 4 & 4 & 4 \\
17 & T22 & Tilt-stretch         & $s_x$
   & .970 & .968 & .966 & .963 & .961 & .959 & .957 & .954 & .952 & .950 \\
18 & T23 & Vignette             & strength
   & .080 & .091 & .102 & .113 & .124 & .136 & .147 & .158 & .169 & .180 \\
19 & T24 & Contrast compress.   & $\alpha$
   & .940 & .933 & .927 & .920 & .913 & .907 & .900 & .893 & .887 & .880 \\
20 & T25 & Non-uniform blur     & $\sigma_{\text{lo/hi}}$
   & .2/.4 & .3/.6 & .3/.9 & .4/1.1 & .5/1.3 & .5/1.6 & .6/1.8 & .7/2.0 & .7/2.3 & .8/2.5 \\
\bottomrule
\end{tabular}

\vspace{1em}

\captionsetup{type=figure}
\centering
\IfFileExists{annexe1cocodegrad.png}{\includegraphics[width=1.2\textwidth,height=0.5\textheight,keepaspectratio]{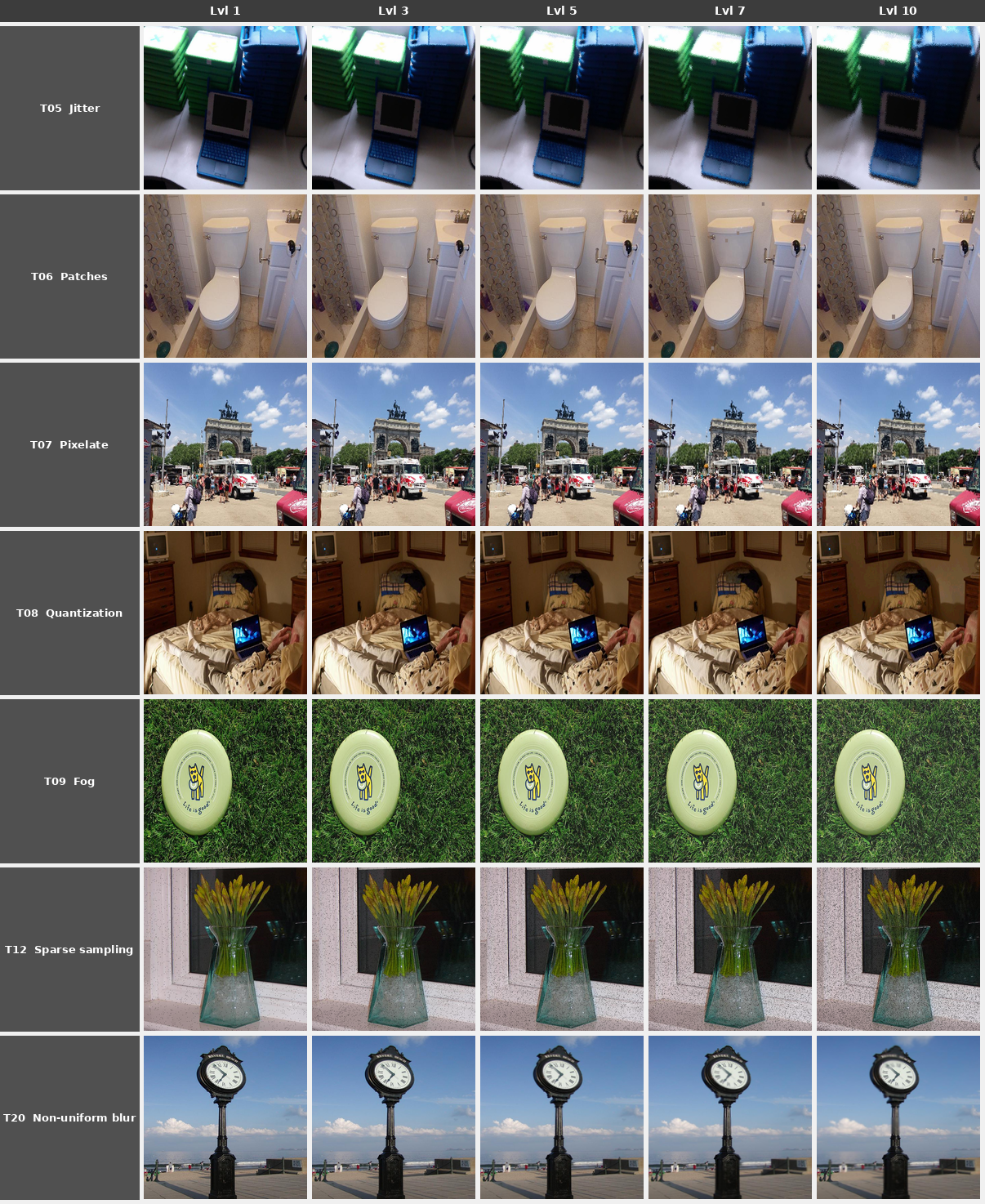}}{\placeholder{5cm}{annexe1cocodegrad.png}}
\caption{Examples of 7 out of 20 degradation types (rows) applied to a MS-COCO reference image at severity levels 1, 3, 5, 7, and 10 (columns).}
\label{fig:appendix_coco}
\end{minipage}

\clearpage
\twocolumn

\vspace{1em}

\begin{figure*}[htbp]
\centering
\IfFileExists{annexe2kadid.png}{\includegraphics[width=\textwidth]{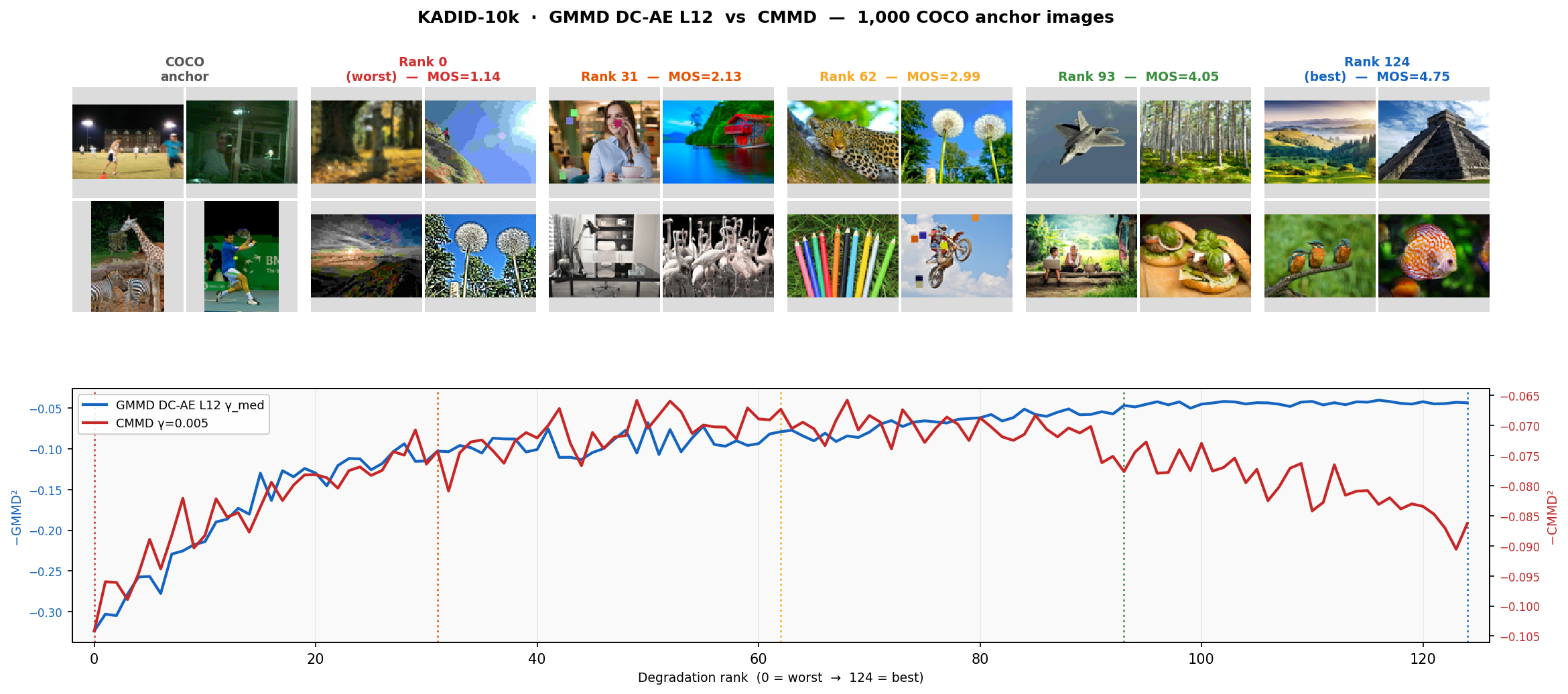}}{\placeholder{8cm}{annexe2kadid.png}}
\caption{KADID-10k evaluation. Top: MS-COCO anchor images (left) and sample images from five degradation groups at ranks 0, 31, 62, 93, and 124 (increasing DMOS). Bottom: \gmmdshort{} and CMMD scores as a function of degradation rank.}
\label{fig:appendix_kadid}
\end{figure*}

\vspace{1.5em}

\begin{figure*}[htbp]
\centering
\IfFileExists{annexe3raise.png}{\includegraphics[width=\textwidth]{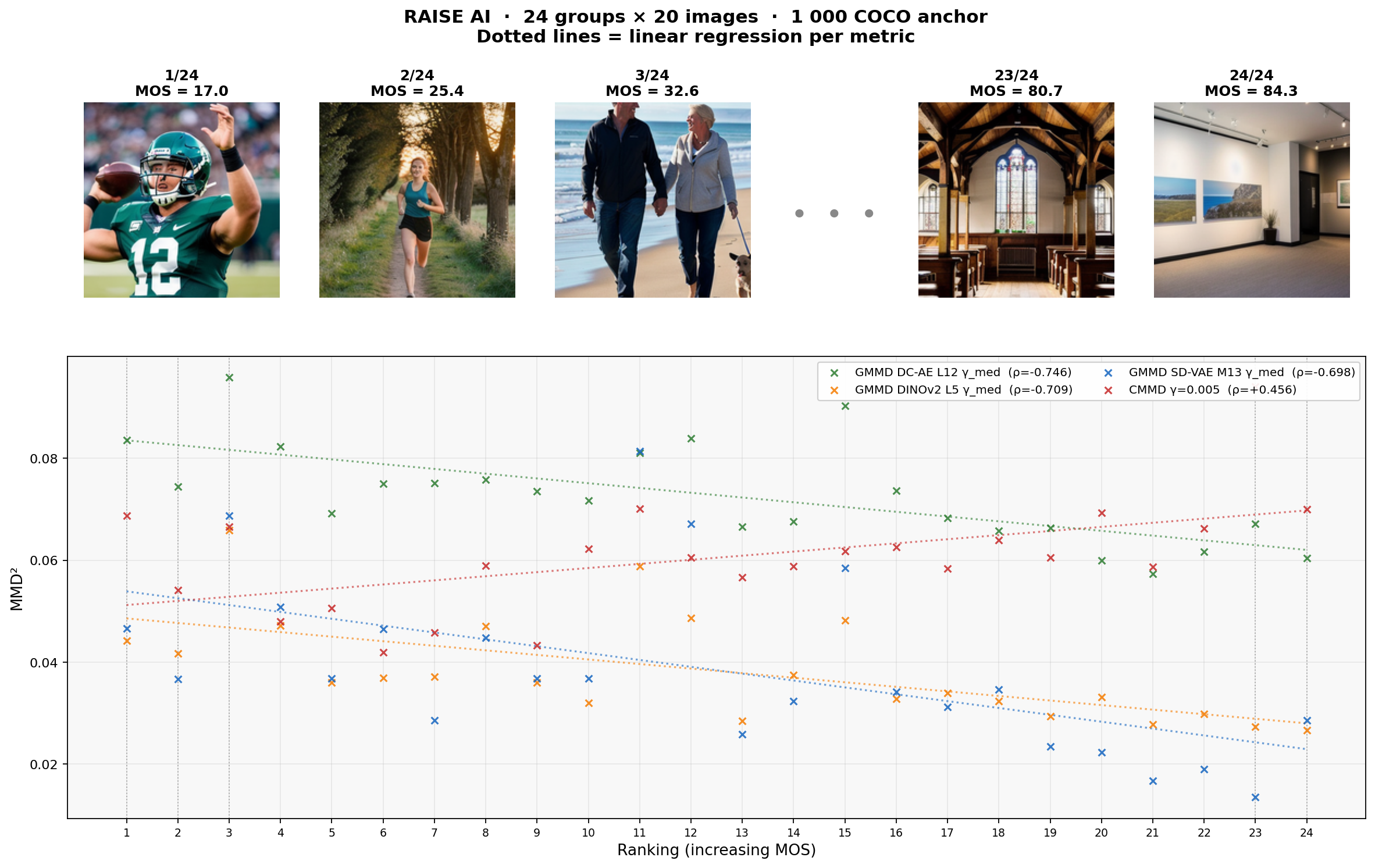}}{\placeholder{8cm}{annexe3raise.png}}
\caption{RAISE realism assessment. Top: AI-generated images from groups at ranks 1, 2, 3, \ldots, 23, 24, sorted by ascending MOS (perceived realism). Bottom: squared MMD vs.\ group rank with linear regression for each \gmmdshort{} configuration and CMMD.}
\label{fig:appendix_raise}
\end{figure*}

\vspace{1.5em}

\begin{figure*}[t]  
\centering
\IfFileExists{annexe4datasets.png}{\includegraphics[width=\textwidth]{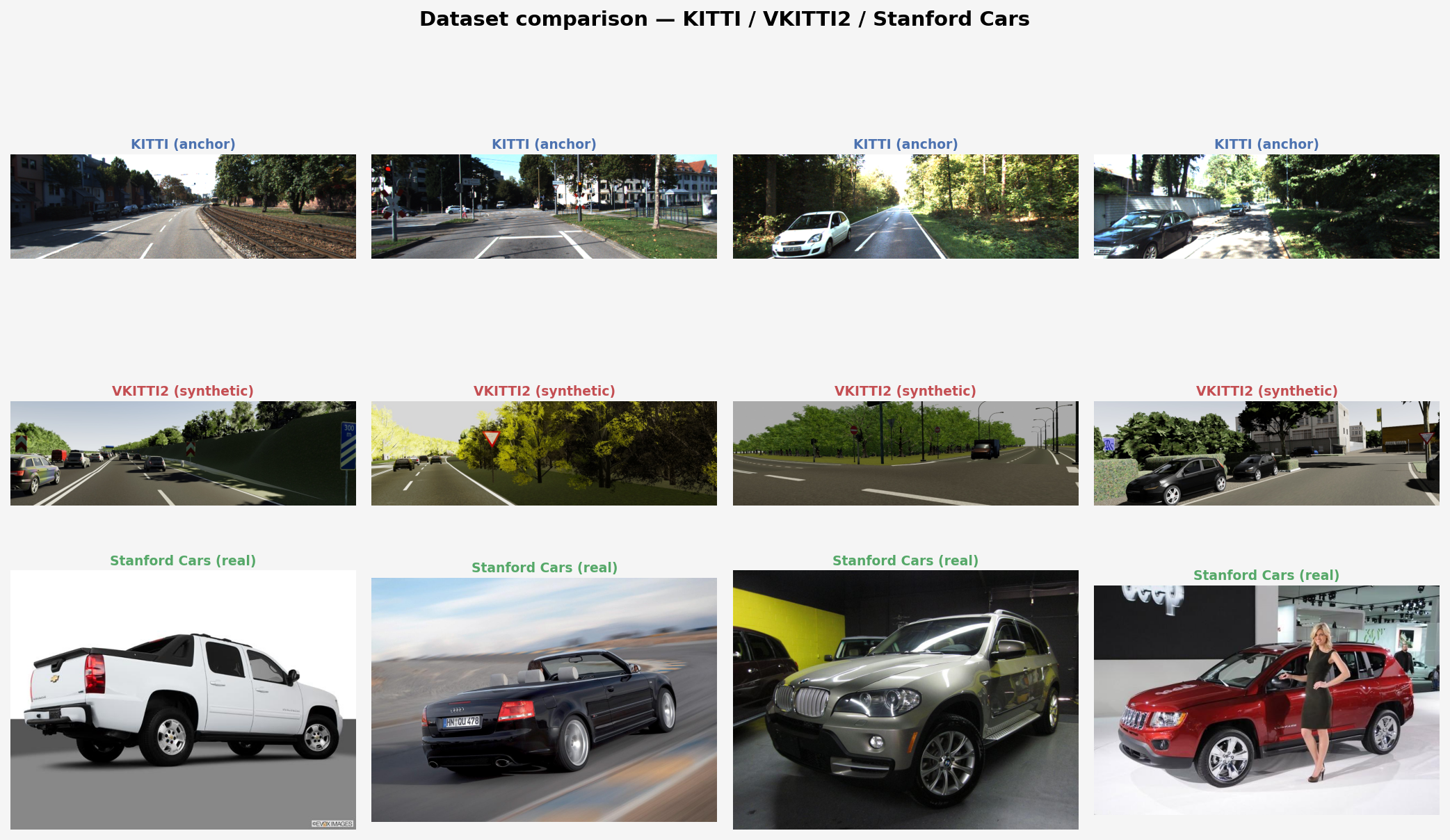}}{\placeholder{6cm}{annexe4datasets.png}}
\caption{Sample images from the three datasets used in the KITTI experiment: KITTI (real driving scenes), Virtual KITTI~2 (synthetic driving scenes), and Stanford Cars (real vehicle photographs).}
\label{fig:appendix_datasets}
\end{figure*}

\end{document}